\documentclass[10pt,twocolumn,letterpaper]{article}

\usepackage{iccv}              

\definecolor{iccvblue}{rgb}{0.21,0.49,0.74}
\usepackage[pagebackref,breaklinks,colorlinks,allcolors=iccvblue]{hyperref}
\usepackage{wrapfig}
\usepackage{graphicx}
\usepackage{amsmath}
\usepackage{amssymb}
\usepackage{booktabs}
\usepackage[dvipsnames]{xcolor}
\usepackage{xcolor,colortbl}
\usepackage{color, colortbl}
\usepackage{xcolor}
\newcommand{\tableCellHeight}{1}
\newcommand{\tabstyle}[1]{
  \setlength{\tabcolsep}{#1}
  \renewcommand{\arraystretch}{\tableCellHeight}
  \centering
  \small
}
\usepackage{sidecap}
\newcommand{\tablestyle}[2]{\setlength{\tabcolsep}{#1}\renewcommand{\arraystretch}{#2}\centering\footnotesize}
\usepackage{float}
\definecolor{purple}{RGB}{230, 227, 254}
\definecolor{lightgreen}{RGB}{238, 252, 241}
\definecolor{lightred}{RGB}{231, 187, 187}
\definecolor{darkred}{RGB}{198, 129, 129}

\definecolor{tabhighlight}{HTML}{e5e5e5}

\newcommand{\rotbox}[1]{\rotatebox{55}{#1}}
\usepackage{graphicx, amsmath, amssymb, caption, subcaption, multirow, overpic}
\definecolor{tabhighlight}{HTML}{e5e5e5}
\definecolor{tabhighlightcolor1}{HTML}{ffe5ec}
\definecolor{tabhighlightcolor2}{HTML}{e3f2fd}
\definecolor{tabhighlightcolor3}{HTML}{d8f3dc}
\definecolor{citecolor}{HTML}{0071bc}

\title{DiMPLe - Disentangled Multi-Modal Prompt Learning: Enhancing Out-Of-Distribution Alignment with Invariant and Spurious Feature Separation}

\author{Umaima Rahman$^1$, Mohammad Yaqub$^1$, Dwarikanath Mahapatra$^2$\\
$^1$Mohamed Bin Zayed University of Artificial Intelligence, $^2$ Khalifa University\\
\\
}

\begin{document}
\maketitle
\begin{abstract}
We introduce \textbf{DiMPLe} (\textbf{Di}sentangled \textbf{M}ulti-Modal \textbf{P}rompt \textbf{Le}arning), a novel approach to disentangle invariant and spurious features across vision and language modalities in multi-modal learning. Spurious correlations in visual data often hinder out-of-distribution (OOD) performance. Unlike prior methods focusing solely on image features, DiMPLe \textbf{disentangles} features \textbf{within and across modalities} while maintaining consistent alignment, enabling better generalization to \textbf{novel classes} and robustness to \textbf{distribution shifts}.
Our method combines three key objectives: (1) mutual information minimization between invariant and spurious features, (2) spurious feature regularization, and (3) contrastive learning on invariant features. Extensive experiments demonstrate DiMPLe demonstrates superior performance compared to CoOp-OOD, when averaged across 11 diverse datasets, and achieves absolute gains of 15.27 in base class accuracy and 44.31 in novel class accuracy. The code will be released publicly upon acceptance.
\end{abstract}    

\section{Introduction}
\label{sec:intro}

Contrastive Language–Image Pretraining (CLIP) \cite{radford2021learning} revolutionized multi-modal learning by linking textual and visual data \cite{jia2021scaling}, \cite{singh2022flava}, \cite{du2022learning}, \cite{gu2021open}, \cite{li2023blip}. The semantic associations present in the shared image-text embedding space allowed it to make predictions on novel and unseen categories, inspiring further research in the domain of multi-modal understanding. 
Prompt tuning methods like CoOp \cite{zhou2022learning} refined CLIP’s performance through context optimization  of learnable prompt vectors, and demonstrate improved text-visual alignment and model adaptability. By employing different prompt-initialization strategies, CoOp successfully showed how prompt engineering could facilitate model adaptability and enable improved alignment between text and visual representations. 
However, despite CoOp’s effectiveness, it exhibited limited generalizability to novel classes. This limitation stemmed from spurious correlations within the visual data, which often misled the model’s predictions when applied to out-of-distribution (OOD) samples. 


\begin{figure*}[ht!]
  \centering
   \includegraphics[width=1.0\linewidth]{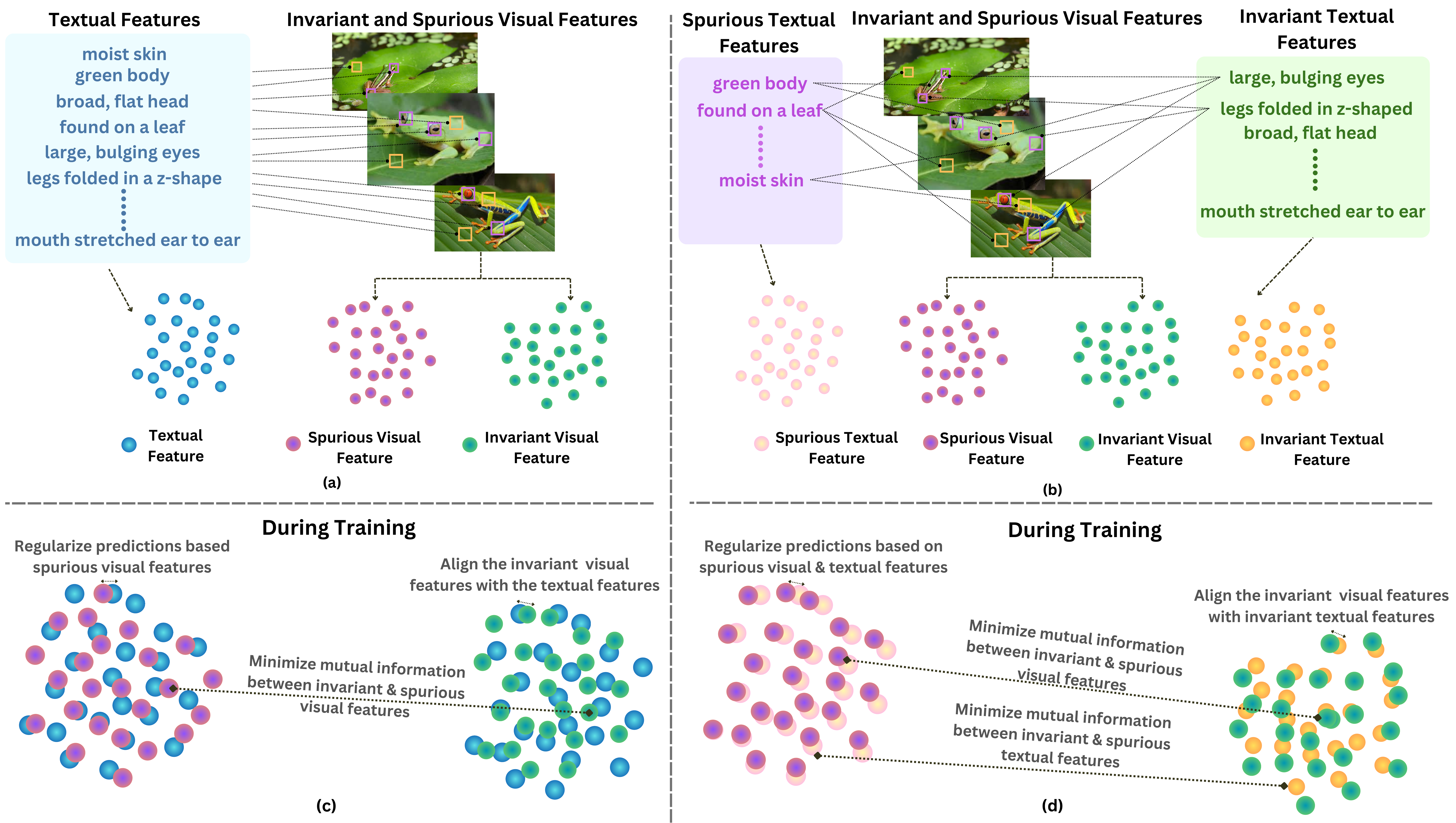}
   \caption{(a) Without the explicit mapping in previous methods, spurious elements in the text embeddings may inadvertently align with the visual model’s invariant features, weakening the model's robustness and reducing its effectiveness on novel classes and out-of-distribution images. (b) Importance of a principled method (\textbf{DiMPLe}) for disentangling multi-modal representations, to enhance generalization for novel classes as well as to improve the robustness of multi-modal models against distribution shifts for the same class by explicitly mapping the invariant and spurious features for each modality. (c) In previous methods, during training, the general textual features are used to align with the invariant image features as well as spurious image features. (d) \textbf{DiMPLe} removes the ambiguous mapping by aligning the invariant textual and image features as well as associating spurious textual features with their spurious image counterparts.}
   \label{fig:concept_Dimple}
   \vspace{-2mm}
\end{figure*}


CoOp-OOD \cite{zhangamend}, an enhancement to the original CoOp, addressed this issue by decoupling invariant and spurious features in the images. By maximizing the similarity between invariant visual features and the corresponding textual representations, it emphasized on robust features, while regulating the predictions based on spurious ones. Additionally, minimizing the conditional mutual information between invariant and spurious image features ensured a clearer separation within the feature space. 

Nonetheless, a crucial consideration in this approach remains unaddressed: spurious features within an image should logically correspond to spurious elements within textual representations, while invariant features should be aligned exclusively with invariant text components. Without explicit mapping, the disentanglement process may lead to ambiguous associations between spurious and invariant features, making it difficult for the model to distinguish relevant signals from noise during inference as seen in Fig. \ref{fig:concept_Dimple} (a) and (c). This ambiguity can propagate through the network, potentially amplifying the influence of spurious correlations under domain shifts or when encountering novel categories. By explicitly aligning spurious and invariant features within and across modalities, the model ensures consistent feature associations, thereby enhancing its interpretability and resilience in applications where distribution shifts are inevitable.

In light of these limitations, we propose \textbf{DiMPLe} - a \textbf{Di}sentangled \textbf{M}ulti-Modal \textbf{P}rompt \textbf{Le}arning a principled approach aimed at separating invariant and spurious features across both vision and language modalities. DiMPLe achieves this by minimizing the conditional mutual information between the invariant and spurious features within each modality. Our contribution can be summarised as: 

\begin{enumerate}\setlength{\itemsep}{1em}
    \item DiMPLe is a coherent approach to feature disentanglement across vision-language modalities avoiding ambiguities across modality mappings (Fig. \ref{fig:concept_Dimple} (b) and (d)).
    \item DiMPLe goes beyond spurious correlation mitigation by employing class-conditioned mutual information minimization that maintains separation between invariant and spurious components across modalities and ensures effective separation of unwanted correlations and associating spurious textual features with its image counterpart.

    \item DiMPLe achieves significant improvements in both base and novel class performance. Ensuring that invariant features remain isolated and consistent, it transfers better across diverse datasets against distribution shifts for the same class, compared to existing methods.

\end{enumerate}

\begin{figure*}[ht!]
  \centering
   \includegraphics[width=0.8\linewidth]{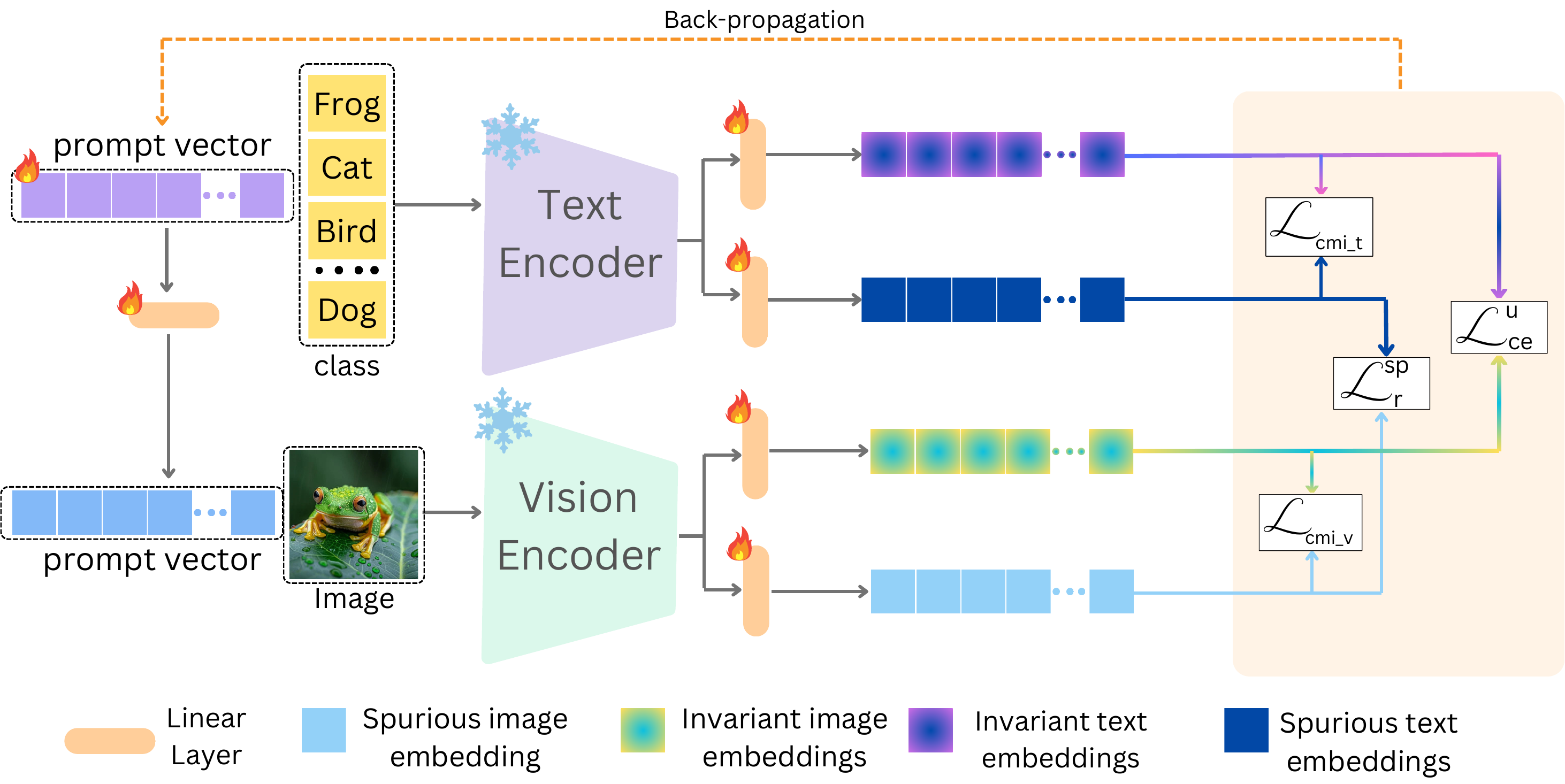}

   \caption{An overview of \textbf{DiMPLe} (\textbf{Di}sentangled \textbf{M}ulti-modal \textbf{P}rompt \textbf{Le}arning), a unified multi-modal prompt-tuning approach that leverages disentangled image and text features, along with multi-stage deep prompting where vision prompts are conditioned on language.}
   \label{fig:DiMPLe}
\end{figure*}

\section{Related Works}
\label{sec:2_rel_works}
Vision-Language Models (VLMs) like CLIP have revolutionized the paradigm of multi-modal training by leveraging the semantic alignment between visual and textual features. The effectiveness of these models heavily depends on how the text prompts are constructed and optimized. Early approaches \cite{jia2021scaling}, \cite{alayrac2022flamingo}, \cite{li2023blip}, relied on hand-crafted prompts, but recent work such as \cite{zhou2022learning}, \cite{jia2022visual}, \cite{zhu2023prompt},  \cite{jin2021good}, \cite{yao2024cpt} have shown that learnable prompts can significantly improve performance. CoOp \cite{zhou2022learning} introduced a paradigm shift by replacing hand-crafted prompts with learnable context vectors while keeping the class names fixed. This approach demonstrated superior performance compared to manual prompt engineering, particularly in base category classification. However, CoOp's performance degraded significantly when applied to novel classes, indicating limitations in its generalization capabilities. MaPLe \cite{khattak2023maple} advanced the field by introducing a meta-learning framework for prompt optimization. The key innovation was the separation of prompts into two components: a shared meta-prompt that captures domain-general knowledge, and class-specific prompts that encode category-specific features. This decomposition allowed for better generalization to unseen classes while maintaining strong performance on base categories. MaPLe's meta-learning strategy enables the model to learn how to adapt prompts quickly for new categories, addressing a key limitation of previous approaches.

Out-of-distribution (OOD) generalization has emerged as a critical challenge for Vision-Language Models, particularly in real-world applications where test distributions may differ significantly from training data. Co-CoOp \cite{zhou2022conditional} specifically addressed the OOD challenge by introducing instance-conditional prompts. Unlike its predecessor CoOp, Co-CoOp generates prompts that are conditioned on the input image, allowing for more flexible and adaptive representations. This approach showed remarkable improvements in OOD scenarios, particularly when dealing with distribution shifts caused by changes in visual styles or environmental conditions. CoOp-OOD \cite{zhangamend} further explored the OOD challenge by introducing a framework specifically designed for improving the generalization of out-of-distribution samples. The method employs a novel training strategy that explicitly considers distribution shifts during prompt optimization. By incorporating both in-distribution and simulated OOD samples during training, CoOp-OOD develops more robust prompt representations that maintain performance across different domains. CLIP-OOD \cite{shu2023clipood} represents another significant advancement in addressing OOD challenges. The work introduces a systematic approach to analyzing and improving CLIP's OOD generalization capabilities. Collectively, these works highlight an important trend in the field: the evolution from static, manually-crafted prompts to increasingly sophisticated, adaptive prompt learning strategies that explicitly consider generalization and robustness. The progression from CoOp to MaPLe and from Co-CoOp to CLIP-OOD demonstrates the field's growing emphasis on developing methods that can handle real-world challenges like distribution shifts and novel class recognition. These developments suggest future research directions, including, integration of multiple prompt learning strategies, investigating techniques for identifying and mitigating specific types of distribution shifts and exploring ways to incorporate domain knowledge into prompt learning frameworks.

\section{Methodology}
\label{sec:3_method_dimple}
Our methodology builds on the foundations laid by CoOp-OOD and further builds on its disentanglement of invariant and spurious image features by adding a disentanglement component for textual features, enhancing robustness to novel classes. Additionally, we incorporate deep vision and language prompts  \cite{khattak2023maple}, which improves feature alignment across modalities. 

\subsection{DiMPLe - An Overview}
Motivated by the synergistic alignment of multiple modalities, it is logical to perform the disentanglement of invariant and spurious features within both the visual and textual feature spaces. As a result, we propose \textbf{DiMPLe} (\textbf{Di}sentangled \textbf{M}ulti-modal \textbf{P}rompt \textbf{Le}arning) (Fig. \ref{fig:DiMPLe}), a unified multi-modal prompt-tuning approach that leverages disentangled image and text features, along with multi-stage deep prompting where vision prompts are conditioned on language, to achieve enhanced OOD generalization.

\textbf{Image Feature Decomposition and Decoupling:}
We start by decomposing image features $\mathbf{z_v}_i$ from the vision encoder into invariant features $\mathbf{z_v}_{i,u}$ captured by a linear projection layer $\phi_v$ and spurious features  $\mathbf{z_v}_{i,s}$  captured by a separate projection layer $\psi_v$. Similarly, we project the textual features  $\mathbf{z_t}_i$ obtained from the textual encoder into invariant features $\mathbf{z_t}_{i,u}$ using a linear projection layer $\phi_t$ and spurious features $\mathbf{z_t}_{i,s}$ using another linear projection layer $\psi_t$. This is formulated as:
     \begin{equation}
     \mathbf{z_v}_{i,u} = \phi_v(\mathbf{z_v}_i), \quad \mathbf{z_v}_{i,s} = \psi_v(\mathbf{z_v}_i)
     \end{equation}
     \begin{equation}
     \mathbf{z_t}_{i,u} = \phi_t(\mathbf{z_t}_i), \quad \mathbf{z_t}_{i,s} = \psi_t(\mathbf{z_t}_i)
     \end{equation}
This decomposition enables us to isolate the stable, class-relevant features from those likely influenced by spurious correlations, thus achieving unbiased alignment in OOD scenarios.

\textbf{Conditional Independence Regularization:}
We enforce a conditional independence constraint between the invariant representations $\mathbf{z}_{i,u}$ and its spurious counterpart $\mathbf{z}_{i,s}$ given the label $Y$ for the vision ($v$) and text ($t$) modalities, using a mutual information regularization term $\mathcal{L}_{cmi}$ to minimize the conditional mutual information $I(\mathbf{z}_{i,s}; \mathbf{z}_{i,s} | Y)$: 
\begin{equation}
\begin{split}
    \mathcal{L}_{cmi\_v} = I(\mathbf{z_v}_{i,u}; \mathbf{z_v}_{i,s} | Y) \text{ for image features,}
    \\
    \mathcal{L}_{cmi\_t} = I(\mathbf{z_t}_{i,u}; \mathbf{z_t}_{i,s} | Y) \text{ for text features,}
    \\
    \mathcal{L}_{cmi} = avg(\mathcal{L}_{cmi\_v}, \mathcal{L}_{cmi\_t}) 
\end{split}
\end{equation}
     
Practically, we estimate $\mathcal{L}_{cmi}$ based on the HSIC metric proposed by \cite{kalinke2023nystrommhilbertschmidtindependencecriterion} which minimizes the conditional mutual information by promoting the independence of variables. It achieves this by computing the Conditional Hilbert-Schmidt Independence Criterion (HSIC) between two sets of features: invariant (x) and spurious (y) conditioned on the class label (z), ensuring that the components capture distinct, relevant information.



\textbf{Deep Multi-modal Prompt Tuning:}
We incorporate multi-modal deep prompting by introducing learnable prompt tokens in both vision and language branches up to the \( J \)-th transformer layer, where \( J < K \), with \( K \) being the total number of layers in the CLIP model.
In the language branch, we introduce \( b \) learnable prompt tokens \( \{P_i \in \mathbb{R}^{d_l}\}_{i=1}^b \), and the prompts are progressively updated within each transformer layer:
     \begin{equation}
     [P_i, W_i] = L_i([P_{i-1}, W_{i-1}])
     \end{equation}
In the vision branch, we add learnable prompts \( \{\tilde{P}_i \in \mathbb{R}^{d_v}\}_{i=1}^b \) that are incorporated at each transformer block, updating the feature representations as:
     \begin{equation}
     [c_i, E_i, \tilde{P}_i] = V_i([c_{i-1}, E_{i-1}, \tilde{P}_{i-1}])
     \end{equation}

\textbf{Vision-Language Prompt Coupling via Linear Projection:}
We incorporate vision-language coupling by conditioning vision prompts \( \tilde{P}_i \) on the language prompts \( P_i \) through a linear projection \( F_i \) that maps text embeddings \( P_i \) into the vision space. This coupling is represented as:
     \begin{equation}
     \tilde{P}_i = F_i(P_i), \quad \tilde{P}_i \in \mathbb{R}^{d_v}
     \end{equation}
This coupling function bridges the vision and language modalities by enabling prompt learning in a shared embedding space, enhancing consistency and complementary representation across both branches.

\textbf{Unbiased Cross-modal Alignment:}
We use a two-fold contrastive learning loss to align the invariant and spurious features in an unbiased manner: (1) We minimize the cross-entropy loss \( \mathcal{L}^{u}_{ce} \) between the invariant image features \( \mathbf{z_v}_{i,u} \) and the text embeddings \( \mathbf{z_t}_{i,u} \):
     \begin{equation}
         \mathcal{L}^{u}_{ce} = - \sum_{i=1}^{N} y_i \log p_u(y_i | x_i)
     \end{equation}
    
    \begin{equation}
    p_u(y_i | x_i) = \frac{\exp(\texttt{sim}(\mathbf{z_v}_{i,u}, \mathbf{z_t}_{i,u}) / \tau)}{\sum_{j=1}^{C} \exp(\texttt{sim}(\mathbf{z_v}_{j,u}, \mathbf{z_t}_{j,u}) / \tau)},
\end{equation}
(2) For the spurious features, we use a uniformity constraint to prevent classification alignment on these features for both modalities, using the Kullback-Leibler (KL) divergence loss \( \mathcal{L}^{sp}_r \):
     \begin{equation}
     \mathcal{L}^{sp}_r = \sum_{i=1}^{N} \ell_{KL}(p_s(y_i | x_i) | p_0)
     \end{equation}
The combined objective function integrates all components:
     \begin{equation}
     \mathcal{L} = \mathcal{L}^{u}_{ce} + \alpha \mathcal{L}^{sp}_r + \beta \mathcal{L}_{cmi}
     \end{equation}
   where \( \alpha \) and \( \beta \) balance the effects of spurious, invariant, and conditional independence terms.

\begin{table}[h!]
    \centering
    \setlength{\tabcolsep}{1mm}{
    \resizebox{1\linewidth}{!}{
    \begin{tabular}{l ccc}
    \toprule
    Method  & Base Acc. & Novel Acc. & HM \\
            \midrule
    \textbf{Independent Vision-Language Prompting}        &  &  &   \\
    1: CoOP-OOD*     & 25.99 & 29.98 &  27.6  \\
    2: DiMPLe*    & \textbf{76.73} & \underline{72.36} &  \underline{74.48} \\
    \midrule
    \textbf{Conditioned Vision-Language Prompting} &  &  &   \\
    1: CoOP-OOD\dag             & 62.72 & 65.6 & 64.13 \\
    \rowcolor{tabhighlightcolor2}
    2: DiMPLe             & \underline{76.09} & \textbf{73.35} &  \textbf{74.70} \\
    \bottomrule
    \end{tabular}
    }}
    \caption{Comparison of DiMPLe with different prompting designs in base-to-novel generalization. Results are averaged over 11 datasets.  IVL-P is independent vision-language prompting, CVL-P is conditioned vision-language prompting where the vision prompts are conditioned on the textual prompts. CoOP-OOD* is CoOP-OOD w/ IVL-P, CoOP-OOD\dag is CoOP-OOD w/ CVL-P and DiMPLe* is DiMPLe using IVL-P. HM is the harmonic mean.
    }
    \label{table:different_V-L_prompting}
    \vspace{-0.25in}
\end{table}
\section{Experiments and Results}
\label{sec:4_experiments}
\noindent


\textbf{Datasets:} For evaluating base-to-novel class generalization and cross-dataset performance, we assess our model on 11 image classification datasets covering a broad range of tasks. These include two general-object datasets, ImageNet \cite{deng2009imagenet} and Caltech101 \cite{fei2004learning}; five fine-grained datasets, OxfordPets \cite{parkhi2012cats}, StanfordCars \cite{krause20133d}, Flowers102 \cite{nilsback2008automated}, Food101 \cite{bossard2014food}, and FGVCAircraft \cite{maji2013fine}; a scene recognition dataset, SUN397 \cite{xiao2010sun}; an action recognition dataset, UCF101 \cite{soomro2012ucf101}; a texture dataset, DTD \cite{cimpoi2014describing}; and a satellite-image dataset, EuroSAT \cite{helber2019eurosat}. For domain generalization, we use ImageNet as the source dataset and test it on four of its variants, including ImageNetV2 \cite{recht2019imagenet}, ImageNetSketch \cite{wang2019learning}, ImageNet-A \cite{hendrycks2021natural}, and ImageNet-R \cite{hendrycks2021many}.

\textbf{Baselines:}
In addition to evaluating DiMPLe against existing  CoOp-OOD \cite{zhangamend} as discussed in the preliminary section, we introduce two more baselines by incorporating vision prompts into the CoOp-OOD pipeline. As discussed in Table \ref{table:different_V-L_prompting}, we specifically create two variants: (1) CoOp-OOD*, which extends CoOp-OOD with vision prompts that operate independently from the language prompts, and (2) CoOp-OOD\dag, which incorporates vision prompts conditioned on the learnable textual prompt vectors. Both CoOp-OOD* and CoOp-OOD\dag are trained using the same loss functions that guide CoOp-OOD, ensuring consistency in optimization objectives across the baselines. 

\textbf{Implementation Details:} 
To implement our approach, we used few-shot learning, selecting 16 random samples per class and fine-tuning a pre-trained ViT-B/16 CLIP model. The model architecture utilized specific dimensions (dl = 512, dv = 768, and dvl = 512) and employed prompt tuning with a depth of 9 layers and prompt lengths of 2 for both language and vision components. The training was conducted on a single NVIDIA A100 GPU over 5 epochs, using an SGD optimizer with a batch size of 4 and a learning rate of 0.0035, with results averaged across 3 runs. We initialized the first layer's prompt vector using CLIP's pre-trained word embeddings for 'a photo of a [category]'. Subsequent layers were randomly initialized using a normal distribution. In cross-dataset evaluation scenarios, we trained on all 1000 ImageNet classes using a reduced prompt depth of 3 layers, training for 2 epochs with a learning rate of 0.0026, measuring performance through base and novel class accuracies and their harmonic mean.


\subsection{Base-to-Novel Class Generalization} 
We evaluate the generalizability of the DiMPLe model in a zero-shot setup, assessing its performance across both base and novel classes to highlight its capability to handle OOD classes effectively. In this setting, each dataset is split into base classes (on which models are trained in a few-shot configuration) and novel classes, which allow us to gauge how well the model can generalize to unseen categories without retraining. Table \ref{table:base2novel} provides a comprehensive comparison of DiMPLe with baseline models, including CLIP \cite{radford2021learning} and CoOp-OOD \cite{zhangamend}. Additionally, we present CoOp-OOD\dag, an enhanced version of CoOp-OOD where we have incorporated conditioned vision prompts to refine its ability to handle novel distributions. Table \ref{table:base2novel} captures performance across various datasets, averaging results for a broad view and presenting specific datasets like ImageNet, Caltech101, and StanfordCars to highlight how each method adapts to new categories. We report three metrics: performance on base classes, performance on novel classes, and the harmonic mean (HM) of the two, which emphasizes the balance between in-distribution (ID) and OOD performance. DiMPLe consistently maintains strong generalization across novel classes compared to both baselines and fine-tuned models. For instance, DiMPLe outperforms CoOp-OOD on novel classes in most datasets, with a notable absolute gain of 15.27 on base class accuracy whereas with 44.31 on the novel class and a harmonic mean of 35.39 on the average of all datasets. This increase is largely due to DiMPLe's disentangling mechanism, which aids in isolating invariant features from spurious ones, enhancing its adaptability to OOD data. Notably, DiMPLe shows more balanced generalization on datasets like Stanford Cars, Flowers102, FGVC-Aircraft, and DTD.

\subsection{ Domain Generalization} 
We examine the robustness of our method by testing it on out-of-distribution datasets. Like in cross-dataset evaluation, we directly evaluate our ImageNet-trained model on four distinct ImageNet variants- ImageNet-A, ImageNet-R, ImageNet-S, and ImageNetV2. These variants represent different types of domain shifts, testing the model's ability to generalize beyond the original ImageNet training data. The results of DiMPLe are compared against CLIP and CoOp-OOD . CLIP represents a general-purpose vision-language model baseline, while CoOp-OOD, is an approach specifically designed for handling OOD shifts, which we anticipate will perform differently from models not specifically optimized for OOD. Table \ref{tab:dg} showcases that across all target datasets, DiMPLe’s balanced scores suggest that it offers consistent performance across different types of domain shifts without compromising in-distribution accuracy.

\begin{table}[h!]
    \centering
\tablestyle{6pt}{1.1}
\addtolength{\tabcolsep}{-4.5pt}
    \begin{tabular}{l ccccc}
    \toprule
    & \textbf{Source} & \multicolumn{4}{c}{\textbf{Target}} \\ \cmidrule(lr){2-2} \cmidrule(lr){3-6}
     & ImageNet & ImageNetV2 & ImageNet-S & ImageNet-A & ImageNet-R \\
    \midrule
    CLIP &  66.73 & 60.83 & 46.15 & 47.77 & 73.96 \\
    \rowcolor{tabhighlightcolor1}
    CoOp-OOD & 38.7 & 34.53 &  24.13 & 25.5  &  27.77 \\
    \midrule
    \rowcolor{tabhighlightcolor2} DiMPLe  & 69.73 & 61.2 & 45.67 & 44.07  &  73.87\\
    \bottomrule
    \end{tabular}
        \caption{ Evaluation under Domain Generalization setting: DiMPLe shows superior performance with respect to CoOp-OOD.
    }
    \label{tab:dg}
    \vspace{-0.12in}
\end{table}
For the in-distribution ImageNet dataset, CLIP achieves a base accuracy of 66.73\%, while DiMPLe attains a close score of 69.73\%, indicating competitive performance in the source domain. In contrast, CoOp-OOD significantly underperforms with an accuracy of 38.7\%, suggesting that CoOp-OOD sacrifices in-distribution performance to handle OOD scenarios. For Out-of-Distribution evaluation, DiMPLe performs similarly to CLIP, while CoOp-OOD struggles significantly across all benchmarks. On ImageNetV2, which introduces minor shifts, DiMPLe (61.2\%) slightly improves over CLIP (60.83\%), while CoOp-OOD struggles at 34.53\%. On ImageNet-S, which assesses robustness to synthetic distortions,  CLIP (46.15\%) and DiMPLe (45.67\%) perform similarly, but CoOp-OOD lags significantly at 24.13\%. For ImageNet-A, which contains adversarially challenging images, CLIP (47.77\%) outperforms DiMPLe (44.07\%), while CoOp-OOD remains the weakest (25.5\%). Finally, on ImageNet-R, which tests adaptation to visual shifts, DiMPLe (73.87\%) remains close to CLIP (73.96\%), whereas CoOp-OOD struggles at 27.77\%. These results highlight that DiMPLe consistently outperforms CoOp-OOD and remains competitive with CLIP, demonstrating stronger domain generalization.

\begin{figure}[h!]
  \centering
   \includegraphics[width=0.8\linewidth]{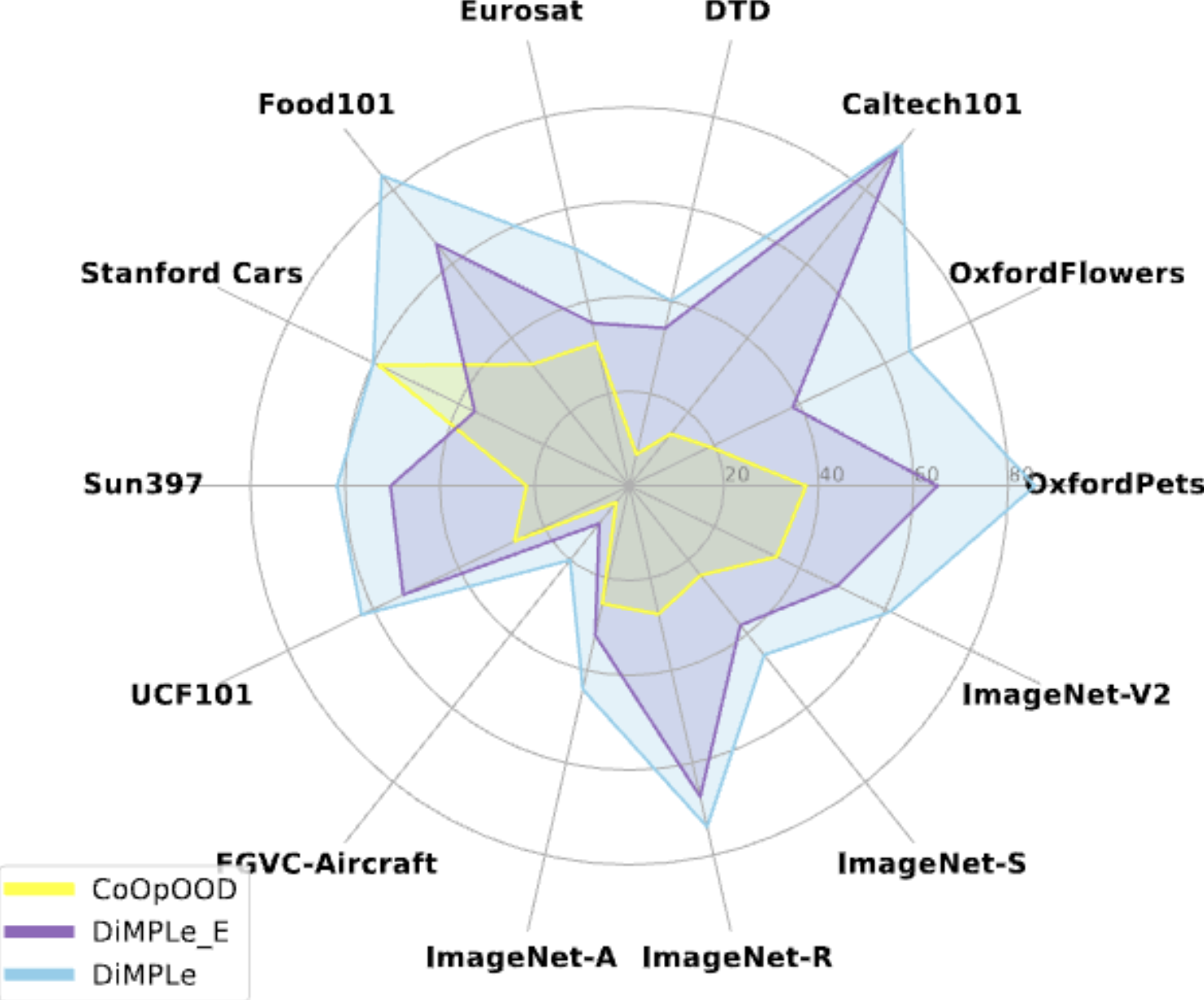}
   \caption{A radar plot showing the results using CoOp-OOD, DiMPLe, DiMPLe-E for Cross-Dataset (11 datasets) and Domain-Generalization (4 ImageNet variants) evaluation. For training, all 1000 classes of ImageNet were used and the respective models were evaluated using the above datasets in a zero-shot setting.}
   \label{fig:radar_plot_xd}
   \vspace{-0.16in}
\end{figure}

\subsection{Cross-Dataset Evaluation} To test the potential for cross-dataset transfer, we apply our ImageNet-trained model to other datasets. Our model undergoes few-shot training on all 1000 ImageNet classes. When comparing CoOp-OOD and DiMPLe across the cross-dataset evaluation, DiMPLe consistently outperforms CoOp-OOD by substantial margins, indicating DiMPLe's robustness in cross-domain generalization. For example, we can see in Fig. \ref{fig:radar_plot_xd} on OxfordPets, DiMPLe achieves 85.67\% accuracy, far surpassing CoOp-OOD’s 37.33\%. This pattern holds across most datasets, with DiMPLe scoring 65.7\% on OxfordFlowers and 92.1\% on Caltech101, whereas CoOp-OOD scores only 18.93\% and 14.07\%, respectively. Notably, DiMPLe also excels on challenging datasets such as ImageNet-A and ImageNet-R, achieving 44.07\% and 73.87\% accuracy, respectively, compared to CoOp-OOD’s 25.5\% and 27.77\%. 

\subsection{Results Discussion}
\noindent


\begin{table*}[t!]
\centering
    \tablestyle{6pt}{0}
    \addtolength{\tabcolsep}{-6pt}
    \tabstyle{1.5pt}
    \setlength{\tabcolsep}{5pt}
    \begin{subtable}[t]{.32\textwidth}
    \centering
    \caption{\textbf{Average over 11 datasets}}
    \begin{tabular}{l cc|c}
    \toprule
    & Base & Novel & HM \\
    \midrule
     CLIP            & 69.34 & \textbf{74.22} & 71.70 \\
     \rowcolor{tabhighlightcolor1}
     CoOp-OOD        & 60.82 & 29.04 & 39.31 \\
     CoOp-OOD\dag    & 62.72 & 65.60 & 64.13\\
    \midrule
    \rowcolor{tabhighlightcolor2}
    \small DiMPLe          & \textbf{76.09} & 73.35 & \textbf{74.70}\\
    \bottomrule
    \end{tabular}
    \end{subtable}
    \vspace{1em}
    \begin{subtable}[t]{.32\textwidth}
    \centering
    \caption{ImageNet.}
    \begin{tabular}{l cc|c}
    \toprule
    & Base & Novel & HM \\
    \midrule
    CLIP & \textbf{72.43} & \textbf{68.14} & \textbf{70.22} \\
    \rowcolor{tabhighlightcolor1}
    CoOp-OOD & 49.13 & 17.03 & 25.29\\
    CoOp-OOD\dag & 62.2 & 55.9 & 59.0\\
    \midrule
    \rowcolor{tabhighlightcolor2}
    DiMPLe & 71.87 & 62.67 &  66.96\\
    \bottomrule
    \end{tabular}
    \end{subtable}
    ~
    \begin{subtable}[t]{.32\textwidth}
    \centering
    \caption{Caltech101}
    \begin{tabular}{l cc|c}
    \toprule
    & Base & Novel & HM \\
    \midrule
    CLIP & 96.84 & \textbf{94.00} & 95.40 \\
    \rowcolor{tabhighlightcolor1}
    CoOp-OOD & 26.1 &  19.2 & 22.12 \\
    CoOp-OOD\dag & 91.2 & 92.57 & 91.88\\
    \midrule
    \rowcolor{tabhighlightcolor2}
    DiMPle & \textbf{97.43} & 93.53  & \textbf{95.44} \\
    \bottomrule
    \end{tabular}
    \end{subtable}
    ~
    \begin{subtable}[t]{.32\textwidth}
    \centering
    \caption{OxfordPets}
    \begin{tabular}{l cc|c}
    \toprule
    & Base & Novel & HM \\
    \midrule
    CLIP & 91.17 & 97.26 & 94.12 \\
    \rowcolor{tabhighlightcolor1}
    CoOp-OOD & 72.97 & 34.73 & 47.06\\
    CoOp-OOD\dag & 85.2  & 96.03 & 90.29\\
    \midrule
    \rowcolor{tabhighlightcolor2}
    DiMPLe & \textbf{91.57} & \textbf{97.73} & \textbf{94.55} \\
    \bottomrule
    \end{tabular}
    \end{subtable}
    \vspace{1em}
    \begin{subtable}[t]{.32\textwidth}
    \centering
    \caption{StanfordCars}
    \begin{tabular}{l cc|c}
    \toprule
    & Base & Novel & HM \\
    \midrule
    CLIP & 63.37 & \textbf{74.89} & 68.65 \\
    \rowcolor{tabhighlightcolor1}
    CoOp-OOD & 76.67 & 69.77 & \textbf{73.06} \\
    CoOp-OOD\dag & 55.83 & 62.23 & 58.86\\
    \midrule
        \rowcolor{tabhighlightcolor2}
    DiMPLe & \textbf{67.97} & 74.07 &  70.89 \\
    \bottomrule
    \end{tabular}
    \end{subtable}
    ~
    \begin{subtable}[t]{.32\textwidth}
    \centering
    \caption{Flowers102}
    \begin{tabular}{l cc|c}
    \toprule
    & Base & Novel & HM \\
    \midrule
    CLIP & 72.08 & \textbf{77.80} & 74.83 \\
    \rowcolor{tabhighlightcolor1}
    CoOp-OOD & 79.13 & 16.17 & 26.85\\
    CoOp-OOD\dag & 58 & 61.93 & 59.9\\
    \midrule
        \rowcolor{tabhighlightcolor2}
    DiMPLe & \textbf{84.33} & 75.00 &  \textbf{79.39}\\
    \bottomrule
    \end{tabular}
    \end{subtable}
    ~
    \begin{subtable}[t]{.32\textwidth}
    \centering
    \caption{Food101}
    \begin{tabular}{l cc|c}
    \toprule
    & Base & Novel & HM \\
    \midrule
    CLIP & 90.10 & 91.22 & 90.66 \\
    \rowcolor{tabhighlightcolor1}
    CoOp-OOD & 74.4 & 34.27 & 46.93\\
    CoOp-OOD\dag & 84.57 & 86.27 & 85.41\\
    \midrule
        \rowcolor{tabhighlightcolor2}
    DiMPLe & \textbf{89.8} & \textbf{91.23} &  \textbf{90.51} \\
    \bottomrule
    \end{tabular}
    \end{subtable}
    \vspace{1em}
    \begin{subtable}[t]{.32\textwidth}
    \centering
    \caption{FGVCAircraft}
    \begin{tabular}{l cc|c}
    \toprule
    & Base & Novel & HM \\
    \midrule
    CLIP & 27.19 & 36.29 & 31.09 \\
    \rowcolor{tabhighlightcolor1}
    CoOp-OOD & 33.27 & 4.83 & 8.44\\
    CoOp-OOD\dag & 16.4 & 23.23 & 19.23\\
    \midrule
        \rowcolor{tabhighlightcolor2}
    DiMPLe & \textbf{30.57} & \textbf{36.8} & \textbf{33.4} \\
    \bottomrule
    \end{tabular}
    \end{subtable}
    ~
    \begin{subtable}[t]{.32\textwidth}
    \centering
    \caption{SUN397}
    \begin{tabular}{l cc|c}
    \toprule
    & Base & Novel & HM \\
    \midrule
    CLIP & 69.36 & \textbf{75.35} & 72.23 \\
    \rowcolor{tabhighlightcolor1}
    CoOp-OOD & 45.23 & 30.03 & 36.1\\
    CoOp-OOD\dag & 71.00 & 66.60 & 68.73\\
    \midrule
        \rowcolor{tabhighlightcolor2}
    DiMPLe & \textbf{75.43} & 74.27 & \textbf{74.85} \\
    \bottomrule
    \end{tabular}
    \end{subtable}
    ~
    \begin{subtable}[t]{.32\textwidth}
    \centering
    \caption{DTD}
    \begin{tabular}{l cc|c}
    \toprule
    & Base & Novel & HM \\
    \midrule
    CLIP & 53.24 & 59.90 & 56.37 \\
    \rowcolor{tabhighlightcolor1}
    CoOp-OOD & 57.2 & 9.53 & 16.34 \\
    CoOp-OOD\dag & 41.7 & 44.23 & 42.93\\
    \midrule
        \rowcolor{tabhighlightcolor2}
    DiMPLe & \textbf{69.8} & \textbf{62.3} & \textbf{65.84} \\
    \bottomrule
    \end{tabular}
    \end{subtable}
    ~
    \begin{subtable}[t]{.32\textwidth}
    \centering
    \caption{EuroSAT}
    \begin{tabular}{l cc|c}
    \toprule
    & Base & Novel & HM \\
    \midrule
    CLIP & 56.48 & \textbf{64.05} & 60.03 \\
    \rowcolor{tabhighlightcolor1}
    CoOp-OOD & 85.7 & 38.63 & 53.25\\
    CoOp-OOD\dag & 55.3 & 63.5 & 59.12\\
    \midrule
        \rowcolor{tabhighlightcolor2}
    DiMPLe & \textbf{80.1}  & 63.93 & \textbf{71.11} \\
    \bottomrule
    \end{tabular}
    \end{subtable}
    ~
    \begin{subtable}[t]{.32\textwidth}
    \centering
    \caption{UCF101}
    \begin{tabular}{l cc|c}
    \toprule
    & Base & Novel & HM \\
    \midrule
    CLIP & 70.53 & \textbf{77.50} & 73.85 \\
    \rowcolor{tabhighlightcolor1}
    CoOp-OOD & 69.17 & 45.3 & 54.75\\
    CoOp-OOD\dag & 68.23 & 69.1 & 68.66\\
    \midrule
        \rowcolor{tabhighlightcolor2}
    DiMPLe & \textbf{78.1} & 75.33 &  \textbf{76.69}\\
    \bottomrule
    \end{tabular}
    \end{subtable}
    ~
    \caption{\small Comparison of our disentangled approach - DiMPLe with CLIP, CoOp-OOD and CoOp-OOD\dag for base-to-novel generalization.}
    \label{table:base2novel}
\end{table*}

\noindent
\textbf{Dataset Size and Sample Diversity Trade-off:}
The impact of dataset size and sample diversity on performance can be critically analyzed by considering how both factors influence model generalization and adaptability across different domains. DiMPLe's performance across varied datasets, ranging from large, well-established datasets like ImageNet to smaller, more diverse datasets like OxfordPets and Stanford Cars, highlights key trends and challenges in model robustness.
As datasets grow in size and complexity, there is an increased risk of overfitting to certain biases in the data (e.g., class imbalance, spurious correlations). DiMPLe mitigates this risk with its disentangling mechanism, which ensures that it learns more invariant and less dataset-specific features, allowing it to generalize better on both large-scale and small, diverse datasets. It benefits from large datasets for in-distribution performance but does not sacrifice its adaptability on smaller, more diverse datasets. 

\begin{figure}[t!]
    \centering
    \begin{subfigure}[b]{\columnwidth}
        \centering
        \includegraphics[width=\columnwidth]{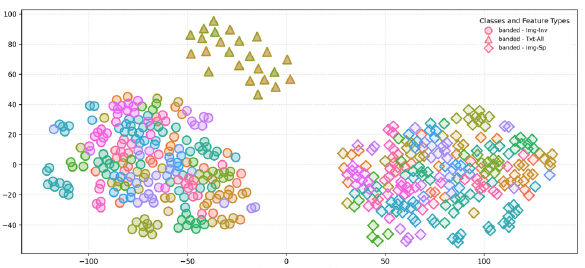}
        \label{fig:tsne_coopood}
    \end{subfigure}
    \vspace{-1mm} 
    \begin{subfigure}[b]{\columnwidth}
        \centering
        \includegraphics[width=\columnwidth]{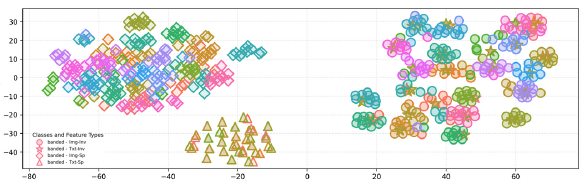}
        \label{fig:tsne_dimple}
    \end{subfigure}
    \vspace{-10mm} 
    \caption{tSNE visualizations showing class discrimination ability in CoOp-OOD (top) compared to DiMPLe (bottom).}
    \label{fig:tsne}
    \vspace{-5mm}
\end{figure}

\noindent
\textbf{Class Discrimination Ability-CoOp-OOD vs DiMPLe:}
The ability to distinctly represent different classes in the feature space, is a crucial aspect of robust image classification, especially in scenarios involving OOD data. We use t-SNE plots to visualize embedding distribution, providing insights into class separation and cluster compactness.
The t-SNE visualizations in Fig. \ref{fig:tsne} compare the the class discrimination capabilities of CoOp-OOD and DiMPLe, revealing notable differences between the two approaches as evident. Firstly, the compactness of clusters in DiMPLe embeddings \ref{fig:tsne} (bottom) indicates lower intra-class variance, meaning that samples from the same class are more tightly grouped. CoOp-OOD embeddings \ref{fig:tsne} (top), however, show looser clusters. Secondly, these visualizations confirm our hypothesis of the importance of segregating the invariant and spurious features for both modalities. Fig. \ref{fig:tsne} (top) presents a tSNE visualization using the CoOp-OOD method. It shows how the invariant image features (circle) are separated from spurious image features (diamond), and all the textual features (triangle) are closer to invariant image features in the embedding space compared to the spurious ones. In contrast to this \ref{fig:tsne} (bottom) presents a tSNE visualization using our approach i.e., DiMPLe. This empirically shows how the textual invariant features (star) are aligned with the image invariant features (circle) whereas the textual spurious features (triangle) which are closer to the image spurious features are well separated from the invariant ones. Thereby involving the multi-modal invariant features to effectively contribute towards the accurate prediction of a class and minimizing the influence of the spurious ones. Thus we can conclude that by explicitly disentangling invariant and spurious features across the vision-language modalities, DiMPLe ensures that invariant embeddings primarily focus on class-relevant information. 

\begin{table*}[t]
    \centering
    \makebox[\textwidth][c]{ 

    \begin{minipage}{0.32\textwidth}
        \centering
        \tablestyle{6pt}{1.1}
        \addtolength{\tabcolsep}{-4.5pt}
        \textbf{Effect of Grouping}
        \vspace{2mm} 
        \begin{tabular}{l|c|c}
            \toprule
            Model (w/o Group Info)  & \multicolumn{2}{c}{\textbf{CelebA}} \\
            & Avg & Worst\\
            \midrule
            CoOp-OOD & 78.10 & 31.11 \\
            \rowcolor{tabhighlightcolor2} 
            DiMPLe  & \textbf{87.36} & \textbf{70.0} \\
            \bottomrule
        \end{tabular}
        \captionsetup{font=small} 
        \caption{Using group information during training. CoOp-OOD vs DiMPLe.}
        \label{table:celeba}
    \end{minipage}
    \hspace{3mm} 

    \begin{minipage}{0.32\textwidth}
        \centering
        \tablestyle{6pt}{1.1}
        \addtolength{\tabcolsep}{-4.5pt}
        \textbf{Effect of Number of Tokens}
        \vspace{2mm} 
        \begin{tabular}{l|cc|cc|cc}
            \toprule
            \# Tokens & \multicolumn{2}{c}{\textbf{Caltech101}} & \multicolumn{2}{c}{\textbf{Food101}} & \multicolumn{2}{c}{\textbf{Flowers102}}\\
            & Base & Novel & Base & Novel & Base & Novel\\
            \midrule
            $n = 1$   & 97.4 & 94.4 & 89.9 & 91.2 & 82.7 & 75.5\\
            \rowcolor{tabhighlightcolor2} 
            $n = 2$   & 97.4 & 93.5 & 89.8 & 91.2 & 84.3 & 75.0\\
            $n = 4$   & 97.4 & 92.8 & 88.6 & 90.3 & 88.0 & 74.6\\
            \bottomrule
        \end{tabular}
        \caption{Varying the number of tokens.}
        \label{table:number_tokens}
    \end{minipage}
    \hspace{3mm} 
    
    \begin{minipage}{0.32\textwidth}
        \centering
        \tablestyle{6pt}{1.1}
        \addtolength{\tabcolsep}{-4.5pt}
        \textbf{Effect of Prompt Depth}
        \vspace{2mm} 
        \begin{tabular}{l|cc|cc|cc}
            \toprule
            Prompt Depth $J$ & \multicolumn{2}{c}{\textbf{Caltech101}} & \multicolumn{2}{c}{\textbf{Food101}} & \multicolumn{2}{c}{\textbf{Flowers102}}\\
            & Base & Novel & Base & Novel & Base & Novel\\
            \midrule
            $J = 3$   & 97.7 & 95.1 & 90.4 & 91.8 & 79.3 & 73.3\\
            $J = 6$   & 97.9 & 92.7 & 90.6 & 91.7 & 83.6 & 75.8\\
            \rowcolor{tabhighlightcolor2} 
            $J = 9$   & 97.4 & 93.5 & 89.8 & 91.2 & 84.3 & 75.0\\
            $J = 12$  & 97.6 & 92.5 & 89.2 & 91.3 & 81.4 & 75.2\\
            \bottomrule
        \end{tabular}
        \caption{Varying the prompt depth $J$.}
        \label{table:prompt_depth}
    \end{minipage}
    }
\end{table*}

\subsection{Ablation Experiments}
\textbf{Optimal points and strategy of disentanglement:}
In order to maximize effectiveness, we must identify the optimal points and strategies for disentangling invariant and spurious features across the two modalities. This leads us to the following question: \textit{how and at what stage should we disentangle the invariant and spurious features across the two modalities?} To answer this question, we analyze an early disentanglement strategy (please see supplementary \ref{supp:method_early}) within the DiMPLe framework which disentangles the features during prompt initialization stage. In this approach, two learnable prompts are designed for each modality and are initialized with templates targeting invariant ('a photo capturing core and invariant features of [category]') and spurious features ('a photo with features that keep changing in [category]'), which are then projected into the visual space using learnable linear layers ($\mu_u$ and $\eta_s$). This design ensures the parallel decomposition of features in both modalities, maintaining alignment through the shared visual-language space. As evident in Fig. \ref{fig:radar_plot_xd} in cross-dataset evaluation, DiMPLe consistently outperforms DiMPLe-E across a variety of datasets, showing better generalization. 

\begin{figure}[t]
    \centering 
    \includegraphics[width=0.8\linewidth]{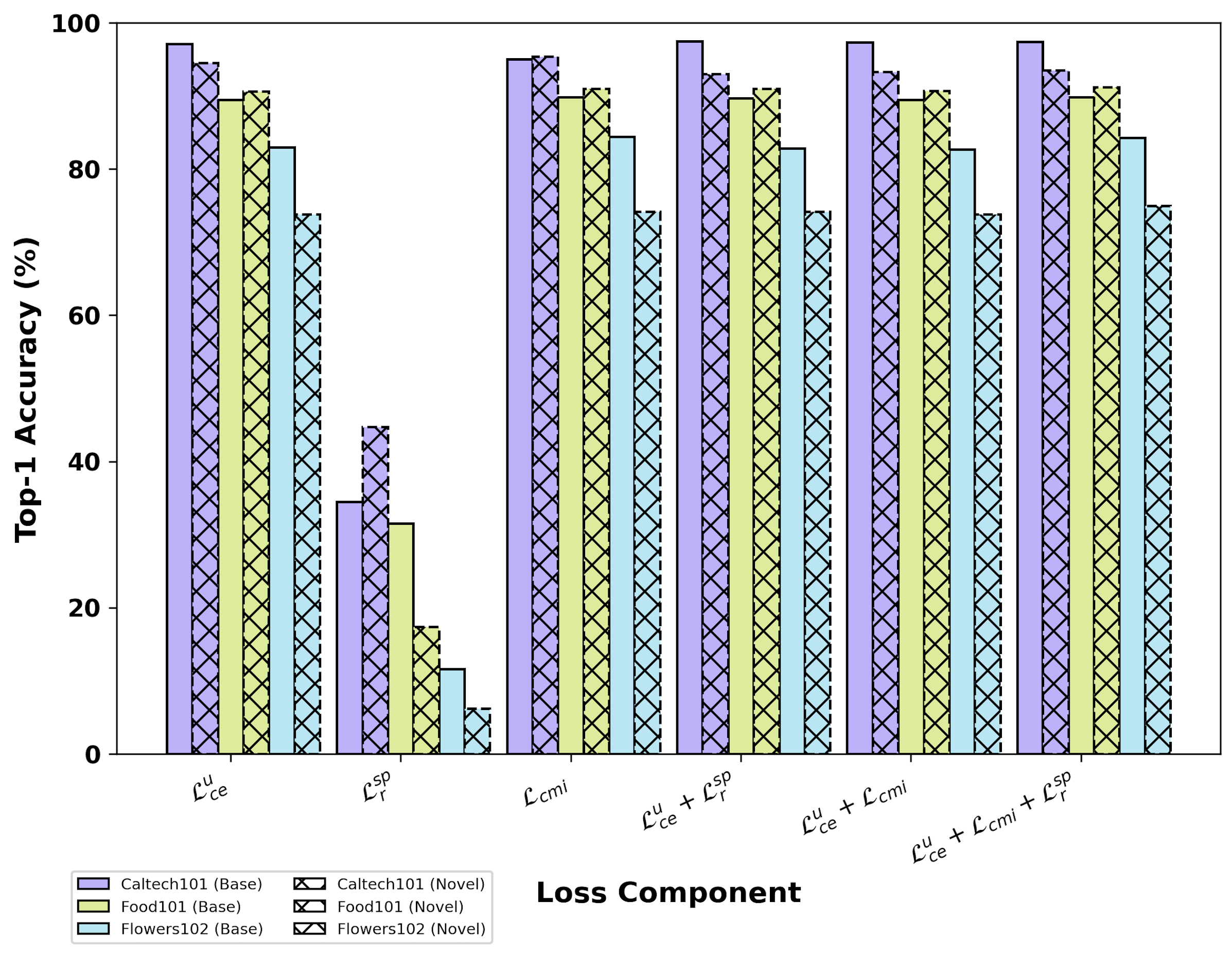}
    \caption{Impact of loss components on model performance.}
    \label{fig:loss_ablation}

    \vspace{-4mm}
\end{figure}

\textbf{Ablation on Loss components:}
Fig. ~\ref{fig:loss_ablation} demonstrates the impact of different loss components on model performance. The baseline model using only the unsupervised cross-entropy loss $\mathcal{L}^{u}_{ce}$ performs well, but adding the class-wise mutual information loss $\mathcal{L}_{cmi}$ further enhances accuracy across datasets. The addition of the spurious feature regularization loss $\mathcal{L}^{sp}_r$ alone leads to suboptimal performance, as seen in the significant accuracy drop, particularly in Food101 and Flowers102.The combination of $\mathcal{L}^{u}_{ce}$ and $\mathcal{L}_{cmi}$ improves robustness while maintaining high accuracy, but our full model, which integrates all three losses, achieves the best overall performance. The inclusion of $\mathcal{L}^{sp}_r$ within the full loss formulation ensures better generalization by reducing reliance on spurious correlations. This is evident from the highest accuracy scores in both base and novel classes across datasets, showing that our approach effectively balances feature disentanglement and task-specific learning without overfitting to spurious features.

\textbf{Effect of Number of Tokens and Prompt Depth}
Tables ~\ref{table:number_tokens} and \ref{table:prompt_depth} examine the impact of varying the number of tokens and prompt depth on model performance respectively, for base-t-novel generalization. For the number of tokens, increasing from $n=1$ to $n=2$ shows minor variations across datasets, with performance remaining stable. However, further increasing to $n=4$ leads to a slight decline in accuracy, particularly for the novel classes in all datasets. This suggests that while additional tokens may enhance feature expressiveness, excessive tokenization could introduce redundancy, leading to diminishing returns. In case of prompt depth, performance trends vary depending on the dataset. Increasing the depth from $J=3$ to $J=6$ results in an improvement for most datasets, with notable gains in Flowers102. However, at $J=9$ and $J=12$, performance either plateaus or slightly declines. This indicates that deeper prompts can enhance representation learning up to a certain extent, beyond which they may lead to overfitting or reduced generalization.

\textbf{Using group information during training:}
Table ~\ref{table:celeba} evaluates model performance on \textbf{CelebA}, a hair color prediction dataset with four groups: 
non-blond females (G1, 3.9\%), non-blond males (G2, 73.9\%), blond females (G3, 21.1\%), and blond males (G4, 1.1\%). The minority group G4 is severely underrepresented, and gender acts as a spurious feature. CoOp-OOD, which optimizes for out-of-distribution (OOD) generalization, achieves an average accuracy of 78.10\%, 
but its worst-group accuracy drops significantly to 31.11\%, showing poor generalization to the minority group. DiMPLe, without explicitly incorporating group information, achieves an improved average accuracy of 87.36\%
and significantly enhances the worst-group accuracy to 70.0\%. This result suggests that DiMPLe generalizes better across groups, mitigating the reliance on spurious correlations and 
achieving robust performance even for underrepresented groups.

\vspace{-2mm}
\section{Conclusion}
\label{sec:5_conclusion}

Our disentangled approach to multi-modal prompt learning DiMPLe, enhances adaptation to out-of-distribution data by isolating invariant and spurious features at both the textual and visual levels. By separating these features, our method improves the alignment of multi-modal information, allowing models to focus on core, class-specific characteristics while filtering out irrelevant correlations. This approach enables robust generalization to unseen categories and distributions by aligning meaningful features across modalities, leveraging feature disentanglement to create a framework better suited for complex, multi-modal tasks.

{
    \small
    \bibliographystyle{ieeenat_fullname}
    \bibliography{main}
}

\clearpage
\setcounter{page}{1}
\maketitlesupplementary
This section contains supplementary material that provides additional details for the main paper and further experimental analysis. This section follows the contents in the following order. 
\begin{itemize}
    \item Methodology - Preliminary Details.
    \item Early disentanglement.
\end{itemize}

\appendix
\section{Methodology - Preliminary Details}
\label{supp:preliminary}
\textbf{Contrastive Language-Image Pre-training (CLIP):}
In the paradigm of multi-modal training, CLIP \cite{radford2021learning} introduces a dual-encoder architecture for the vision and text modalities. Extensive pair of images and its associated caption is used to train these encoders. The caption encoded by the textual encoder $f$ is mapped to a textual feature space whereas its associated image encoded by the vision encoder $g$ is mapped on to a visual feature space. The ultimate goal is to align the image and text features belonging to a corresponding pair of images whereas repel other textual and visual features away through contrastive learning approach. For the downstream task of image classification, CLIP can be employed in a prompt-based approach. In zero-shot classification, rather than undergoing additional training on the specific target classes (e.g., "dog" or "car") within the dataset, CLIP leverages hand-crafted text prompts such as ``\texttt{a photo of a [CLASS]}", pass it through the textual encoder and obtain the textual embeddings corresponding to $c$ classes where $c \in \{1,..., C\}$. Given a new image $x_{test}$, CLIP computes its prediction probability $p(c|x_{test})$ using the similarity between the image embeddings and embeddings of each class prompt, effectively identifying which class label best aligns with the image based on pre-existing learned associations. This process enables CLIP to perform classification without direct exposure to task-specific data, facilitating adaptability across various classification tasks.

\begin{equation}
\vspace{-1mm}
    p(c|x_{test}) = \frac{\exp(\texttt{sim}(\mathbf{z}_{test},\mathbf{w}_c)/\tau)}{\sum_{j=1}^C \exp(\texttt{sim}(\mathbf{z}_{test},\mathbf{w}_j/\tau)}
\vspace{-1mm}
\end{equation}

\noindent
\newline
\textbf{Context-Optimization for Language modality (CoOp):}
Owing to the tedious task of manual prompt-engineering to achieve optimal performance using the CLIP model, authors in \cite{zhou2022learning} introduced a sophisticated methodology of using soft prompts instead of the hand-crafted ones to describe an image. These soft prompts are learnable textual vector denoted as $\mathbf{v} = [v_1, v_2, ..., v_L]$. Each $v_l, l \in {1, ..., L}$ is a vector of the same dimension as the word embeddings and L is a hyper-parameter which controls the number of context tokens. Hence, CoOp introduces a soft (learnable) prompt represented as:
\begin{equation}
    \vspace{-1mm}
    \mathbf{t} = [v_1, v_2, ..., v_L, 
    \texttt{CLASS}]
    \vspace{-1mm}
\end{equation}
For the downstream task of image classification, CoOp adapts CLIP for few-shot transfer by fine-tuning the above continuous set of prompt vectors within its language component. For a given set of $N$ training images $\mathcal{D}_{train} = {(x_i, y_i)}^{N}_{i=1}$, CoOp computes the class probabilities and minimizes the cross-entropy loss $\mathcal{L}_{ce}$ to tune the learnable prompt vector. This optimization problem is expressed as:
\begin{equation}
    \vspace{-1mm}
    t^* = \arg \min_t \mathcal{L}_{\text{ce}},
    \vspace{-1mm}
\end{equation}

\begin{equation}
    \vspace{-1mm}
    \mathcal{L}_{\text{ce}} = -\sum_{i=1}^{N} y_i \log p(y_i | x_i),
    \vspace{-1mm}
\end{equation}

\begin{equation}
    \vspace{-1mm}
    p(y_i | x_i) = \frac{\exp(\texttt{sim}(\mathbf{z}_i, \mathbf{\tilde w}_{y_i}) / \tau)}{\sum_{j=1}^{C} \exp(\texttt{sim}(\mathbf{z}_i, \mathbf{\tilde w}_j) / \tau)},
\end{equation}

with $\mathbf{z}_i = f(x_i)$ are the image embeddings for image $x_i$ and $\mathbf{\tilde w}_{y_i} = g(\mathbf{t}_{y_i}) = g([\mathbf{v}, y_i])$ are the learned textual embeddings. Here $\mathcal{L}_{ce}$ is computed using the similarity between image and text embeddings $\mathbf{z}_i$, $\mathbf{\tilde w}_{y_i}$ respectively. 

\noindent
\newline
\textbf{Context-Optimization with Decoupled Image Features (CoOp-OOD):}
One of the shortcomings of the CoOp approach is its poor generalization capabilities when faced with novel classes. The reason being learning of spurious correlations during the fine-tuning stage. Moreover, using features extracted from the frozen encoders make CoOp vulnerable to bias and inaccurate learning. To address this, authors in \cite{zhou2022learning} propose CoOp-OOD which decouples the image features into invariant and spurious components through two projection layers, $\phi$ and $\psi$. These layers separate the features without additional encoders. To ensure effective decoupling, they employ a structured causal model that describes the relationships between variables using conditional independence to minimize the mutual information between the invariant and spurious image features. 

\noindent
\newline
\textbf{Multi-Modal Prompt Learning (MaPLe):}
The concept of introducing learnable prompt vectors instead of hard-coded ones was an important step in the direction of prompt optimization. However, the generalization performance of CoOp dropped significantly on novel classes and as a result the authors evolved their learning strategy by conditioning these soft prompts on image instances during the fine-tuning stage. They called this method conditional context optimization (Co-CoOp) \cite{zhou2022conditional}. Inspired by this work, authors in \cite{khattak2023maple} designed a Multi-Modal Prompt Learning (MaPLe) framework, which brings in multi-modal learnable prompt vectors instead of using conditioned soft prompts for the text modality alone. 
The language branch of CLIP consists of $b$ learnable tokens $P_i$ concatenated with fixed tokens $W_0$ corresponding to the \texttt{[CLASS]} embedding which  gets processed through $J$ transformer layers of the textual encoder. Similarly, the vision branch also consists of $b$ learnable tokens $\tilde P_i$ concatenated with the input image tokens $E_i$. These tokens are processed through the transformer layers of the image encoder up to depth $J$. The multi-modal prompts are propagated through the associated transformer layers of both textual and visual encoders respectively, in order to achieve better generalization capability. Furthermore, MaPLe employs ``branch-aware" prompt coupling, where vision prompts $\mathbf{\tilde{P}}$ are generated as linear projections of language prompts $\mathbf{P}$ using a function $\mathcal{F}(\cdot)$. This coupling enables mutual gradient propagation and alignment between the visual and textual features, improving performance.

\section{Early disentanglement}
\label{supp:method_early}
To disentangle the invariant and spurious features across vision-language modalities before the encoding stage, we incorporate the following strategy:
\\
\textbf{Feature Decomposition via Early Prompting:}
We introduce two distinct learnable prompts for each modality (vision and language): one for the invariant features and another for the spurious features. These prompts are designed to capture and decouple the task-relevant information (invariant) from the noise or spurious correlations. For each modality, two prompts are introduced: (1) Invariant Textual Prompt (\( P_{x,u} \)): Initialized with a template focused on invariant features, e.g., ``\texttt{a photo capturing core and invariant features of a [class]}". (2) Spurious Textual Prompt (\( P_{x,s} \)): Initialized with a template designed to capture spurious features, e.g., ``\texttt{a photo with features that keep changing of a [class]}". Each prompt is a learnable vector that will evolve through the training process.

\begin{figure}[h]
  \centering
   \includegraphics[width=1.0\linewidth]{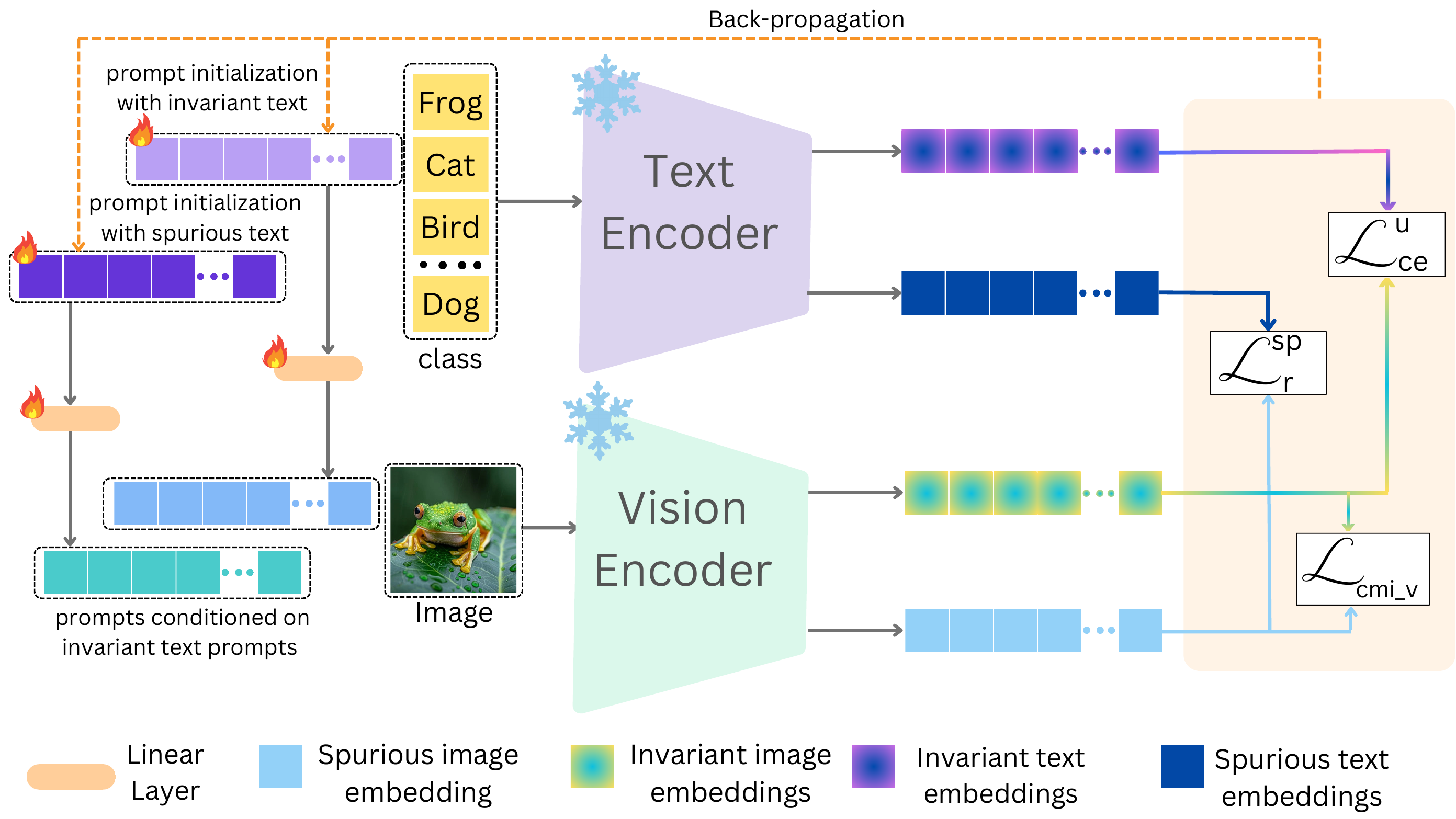}
   \caption{An overview of early disentanglement using separate prompt initializations to disentangle the features before encoding.}
   \label{fig:early}
   \vspace{-3mm}
\end{figure} 

\noindent
\textbf{Projection of Textual Prompts to Visual Prompts:}
Once the invariant and spurious prompts for the text modality are initialized, they are projected into the visual space to influence the corresponding visual representations. This is achieved through a linear projection layer for both prompts: (1) The invariant textual prompt vector \( P_{x,u} \) is projected into the visual space via a learnable linear layer \( \mu_{u} \), resulting in an invariant visual prompt \( \tilde{P}_{x,u} \):
  \begin{equation}
      \tilde{P}_{x,u} = \mu_{u}(P_{x,u})
  \end{equation}
Similarly, the spurious textual prompt vector \( P_{x,s} \) is projected into the visual space using another learnable linear layer \( \eta_{s} \), resulting in a spurious visual prompt \( \tilde{P}_{x,s} \):
  \begin{equation}
  \tilde{P}_{x,s} = \eta_{s}(P_{x,s})
  \end{equation}
These projections allow for the disentanglement of the visual features in the shared visual-language space by guiding the vision branch’s understanding of invariant and spurious components separately.

\noindent
\textbf{Disentangling Vision and Language Features:}
The disentanglement process happens early in the pipeline, before the cross-modal interaction, by ensuring that the vision prompts \( \tilde{P}_{x,u} \) and \( \tilde{P}_{x,s} \) guide the visual encoder in capturing invariant and spurious features, respectively. Similarly, the language encoder processes the invariant and spurious features separately through the respective prompt vectors \( {P}_{x,u} \) and \( {P}_{x,s} \). Let the text encoder output embeddings for the invariant and spurious prompts as \( T_{x,u} \) and \( T_{x,s} \), respectively, and the visual encoder outputs \( V_{x,u} \) and \( V_{x,s} \). The disentanglement occurs when these separate prompts influence the final embeddings produced by the vision and language encoders.
The use of separate prompts for invariant and spurious features allows the model to disentangle the vision and language representations effectively. The vision encoder is guided to capture only the invariant features using the invariant prompts, while the spurious features are isolated using the spurious prompts. The same process applies to the language encoder, where the invariant and spurious prompts guide the encoding process at an early stage, ensuring that the invariant and spurious features are disentangled before the final multi-modal encoding.


\begin{SCtable*}[][t!]
    \tabstyle{4pt}
    \scalebox{0.85}{
    \begin{tabular}{l c cccccccccc|c}
    \toprule
    & \rotbox{\textbf{ImageNet}} & \rotbox{\textbf{Caltech101}} & \rotbox{\textbf{OxfordPets}} & \rotbox{\textbf{StanfordCars}} & \rotbox{\textbf{Flowers102}} & \rotbox{\textbf{Food101}} & \rotbox{\textbf{Aircraft}} & \rotbox{\textbf{SUN397}} & \rotbox{\textbf{DTD}} & \rotbox{\textbf{EuroSAT}} & \rotbox{\textbf{UCF101}} & \rotbox{\emph{Average}} \\
    \midrule
    Base    & 66.43 & 97.57 & 95.10 & 66.07 & 92.10 & 87.20 & 29.07 & 74.23 & 78.33 & 76.70 & 76.93 & 76.34 \\
    Novel   & 62.80 & 92.93 & 95.27 & 69.80 & 68.70 & 89.80 & 27.43 & 69.73 & 50.90 & 70.23 & 69.63 & 69.75 \\
    \midrule
    \rowcolor{tabhighlightcolor2} 
    \textbf{HM}      & 64.56 & 95.19 & 95.18 & 67.88 & 78.70 & 88.48 & 28.23 & 71.91 & 61.70 & 73.32 & 73.1 & 72.57\\
    \bottomrule
    \end{tabular}}
        \caption{Base-to-Novel Class Generalization using the early disentanglement method to separate the invariant and spurious features. The hyper-parameters were same as our main experiments.}
    \label{tab:b2n_early_dimple}
\end{SCtable*}

\noindent
\textbf{Objective Function for Invariant \& Spurious Features:}
Similar to the Late DiMPLe approach in order to ensure the learned representations align as intended, we introduce a triple-objective function: 
\\
(1) Invariant Alignment Loss (\( \mathcal{L}^u_{ce} \)): This loss minimizes the cross-entropy between the invariant visual features \( V_{x,u} \) and the invariant textual embeddings \( T_{x,u} \):
  \begin{equation}
  \mathcal{L}^u_{ce} = - \sum_{i=1}^{N} y_i \log p_u(y_i | V_{x,u}, T_{x,u})
  \end{equation}
(2) Spurious Unalignment Loss (\( \mathcal{L}^{sp}_r \)): A regularization term is applied to ensure the spurious features do not contribute to the classification. This is achieved by a uniformity constraint, ensuring the spurious components \( V_{x,s} \) and \( T_{x,s} \) do not align effectively with the target label:
  \begin{equation}
  \mathcal{L}^{sp}_r  = \sum_{i=1}^{N} \ell_{KL}(p_s(y_i | V_{x,s}, T_{x,s}) | p_0)
  \end{equation}
(3) Conditional Mutual Information Loss $\mathcal{L}_{cmi\_v}$: In Early DiMPLe, we enforce a conditional independence constraint between the invariant and spurious features for the vision modality alone. Specifically, we aim to minimize the conditional mutual information between the invariant and spurious components, conditioned on the label \( Y \), using a mutual information regularization term \( \mathcal{L}_{cmi_v} \). This is achieved through:

\begin{equation}
\mathcal{L}_{cmi\_v} = I(V_{x,i}, V_{x,s}| Y) \quad 
\end{equation}

The total objective function for Early DiMPLe is then:
\begin{equation}
\mathcal{L} = \mathcal{L}^u_{ce} + \alpha \mathcal{L}^{sp}_r + \beta \mathcal{L}_{cmi\_v}
\end{equation}
where \( \alpha \) is a hyperparameter balancing the influence of the invariant and spurious loss terms.

The key difference in Early DiMPLe is that the image and text features are disentangled into invariant and spurious components at an early stage in the pipeline. As a result, the alignment loss is not calculated between the original image and text embeddings, but rather between their invariant components. This is why the loss function uses the invariant features \( V_{x,u} \) and \( T_{x,u} \) for the alignment, instead of the original image \( x_i \) and label \( y_i \).
\\

\textbf{Base-to-Novel Class Generalization for early disentanglement.}
We evaluate the generalizability of the Early DiMPLe approach in a zero-shot setup, assessing its performance across both base and novel classes to highlight its capability to handle out-of-distribution (OOD) classes effectively. In this setting, each dataset is split into base classes (on which models are trained in a few-shot configuration) and novel classes, allowing us to measure how well the model can generalize to unseen categories without additional retraining. Table \ref{tab:b2n_early_dimple} provides a detailed analysis of Early DiMPLe's performance, focusing on its ability to separate invariant and spurious features before encoding, thereby enhancing generalization. From the results, it is evident that Early DiMPLe achieves high Base accuracy across datasets such as Caltech101 (97.57\%) and OxfordPets (95.10\%), demonstrating strong alignment with the base distribution. Similarly, the method shows competitive Novel accuracy, particularly on OxfordPets (95.27\%) and Food101 (89.80\%), highlighting its capacity to generalize effectively to unseen categories. The HM scores provide a more balanced perspective, with an average of 72.57\% across datasets. Notable strengths are observed on datasets like Caltech101 (95.19\%) and Flowers102 (78.70\%), where the model maintains a favorable trade-off. However, challenges remain for datasets such as Aircraft and DTD, where lower HM values indicate difficulties in handling more complex or domain-specific novel distributions. This underscores potential areas for further refinement, such as incorporating more specialized backbones or advanced loss functions tailored to these datasets.  

\noindent
\textbf{Computational complexity for Early vs late disentanglement:}
The early and late disentanglement in the DiMPLe framework reflects key differences in computational complexity. DiMPLe-E, with its early disentanglement approach, has a higher number of parameters and GFLOPs compared to DiMPLe, as mentioned in Table \ref{tab:computationalcomparison}. Despite having a lesser number of trainable parameters and requiring comparatively lesser computations, DiMPLe's performance is better across all three evaluation settings: (1) Base-to-Novel Generalization, (2) Cross-dataset evaluation and (3) Domain Generalization. 

\begin{table}[!h]
\tablestyle{8pt}{1.1}
\addtolength{\tabcolsep}{-3.7pt}
\scalebox{01}{
\begin{tabular}{lcccc}
\toprule
    {Method} & {\# Parameters} & {Parameters \% CLIP} & GFLOPs \\ 
    \midrule
    \rowcolor{tabhighlightcolor2}
    DiMPLe       &  4.08M & 3.28 & 1833.4  &\\
    \rowcolor{tabhighlightcolor3}
    DiMPLe-E     &  6.72M & 5.39 & 1903.8  &\\
\bottomrule
    \end{tabular}
} 
\caption[caption]{{Computational complexity for late disentanglement (DiMPLe) vs early disentanglement (DiMPLe-E). We can see that DiMPLe is a better choice in terms of fewer GFLOPs and fewer tunable parameters compared to DiMPLe-E.}
\label{tab:computationalcomparison}
}
\vspace{-0.2in}
\end{table}
\end{document}